\documentclass[10pt]{article}
\usepackage{colacl}
\usepackage{epsfig}
\usepackage{amssymb}
\newcommand{\epsfscaledbox}[2]{\centerline{\psfig{figure=#1,width=#2}}}
\newcommand{\qset}[2]{\{#1 ~ | ~ #2\}} 
\newcommand{\lsi}{LSI}
\newcommand{\mv}[1]{\mathbf #1} 
\newcommand{\mAhat}{\mv{D}} 
\newcommand{\vAhat}{\mv{d}} 
\newcommand{\numdocs}{n}
\newcommand{\numtopics}{k}
\newcommand{\collection}{{\cal C}}
\newcommand{\colltopics}{{\rm topics}(\collection)} 
\newcommand{\rel}[2]{\mbox{rel}(#1,#2)} 
\newcommand{\simmatrix}{\mv{S}}
\newcommand{\conting}{C}
\newcommand{\assoc}{{\cal N}}
\newcommand{\high}{ceiling}
\newcommand{\low}{floor}
\newcommand{\revalx}{s} 
\newcommand{\asfd}{{\sc auto-scale}}
\newcommand{\sTsim}{\mu(\collection)}
\newcommand{\pop}[1]{\Delta_#1}
\newcommand{\popMin}{\Delta_{min}} 
\newcommand{\popMax}{\Delta_{max}}
\newcommand{\fError}{{\mbox{diff}_{\simmatrix,\mAhat}}} 
\newcommand{\mBase}{\mv{B}}      
 
\newcommand{\mProj}[1]{\mv{P}_{#1}}  
  
\newcommand{\sOptErr}{{\epsilon_{opt}}}  
\newcommand{\sOrgErr}{\epsilon_{vsm}} 
\newcommand{\sOrgErrapprox}{\widehat{\epsilon}_{vsm}} 
\newcommand{\nonuniapprox}{f}
\newcommand{\svdopt}{g}
\newcommand{\oursopt}{\widehat{g}}
\newcommand{\ours}{\mbox{IRR}}
\newcommand{\vsm}{{\rm VSM}}
\newcommand{\Range}[1]{{\rm range}(#1)}
\newcommand{\mA}{\widehat{\mv{D}}}

\newcommand{\sv}{\tau}	 
\newcommand{\svMax}{\widehat{\Delta}_{max}}
\newcommand{\svMin}{\widehat{\Delta}_{min}}

\newcommand{\mAbar}{\overline{\mv{D}}}

\newcommand{\normF}[1]{|| #1 ||_F}
\newcommand{\nTwo}[1]{|| #1 ||_2} 
\newtheorem{theorem}{Theorem}[section]
\newcommand{\Lsi}{LSI}
\newcommand{\irrdim}{\ell}
\newcommand{\makematrix}[1]{\left[\matrix{#1}\right]}
\newcommand{\spaceAny}{{\cal X}}
\newcommand{\spaceOpt}{{\cal X}_{opt}}
\newcommand{\spaceOrg}{{\cal X}_{VSM}}
\newcommand{\spaceLSI}{{\cal X}_{LSI}}
\newcommand{\oProj}[1]{\mv{P_{#1}}}     

\newcommand{\projj}[1]{{\rm proj}^{(#1)}}
\newcommand{\residj}[2]{\mv{r}_{#1}^{(#2)}}
\newcommand{\irrscale}[2]{{\rm pow}(#1, #2)} 
\newcommand{\mvapr}[1]{\widehat{\mv{#1}}}
\newcommand{\precn}{{\rm prec}}
\newcommand{\stewmatrix}{\mv{A}}
\newcommand\qed{ }

\begin{document}
\author{Rie Kubota Ando and Lillian Lee  \\
        Department of Computer Science \\ Cornell University \\
        Ithaca, NY 14853-7501  \\
        {\tt \{kubotar,llee\}@cs.cornell.edu} }
\title{\vspace{-55pt}
{\normalsize \tt \hfill To appear in the proceedings of SIGIR 2001} \\
        \mbox{}\\Iterative Residual Rescaling: \\  An Analysis and Generalization of LSI}
\date{}
\maketitle
\begin{abstract}
We consider the problem of creating document representations in which
inter-document similarity measurements correspond to semantic
similarity.  We first present a novel {\em subspace-based} framework
for formalizing this task. Using this framework, we derive a new
analysis of {\em Latent Semantic Indexing} (LSI), showing a precise
relationship between its performance and the {\em uniformity} of the
underlying distribution of documents over topics.  This analysis helps
explain the improvements gained by Ando's (2000) {\em Iterative
Residual Rescaling} (\ours) algorithm: \ours\ can compensate for
distributional non-uniformity.  A further benefit of our framework is
that it provides a well-motivated, effective method for automatically
determining the rescaling factor \ours\ depends on, leading to further
improvements.  A series of experiments over various settings and with
several evaluation metrics validates our claims.
\end{abstract}

\section{Introduction}
\label{sec:introduction}

\paragraph{Background}  The rapid increase in the availability of
electronic documents has created high demand for automated text
analysis technologies such as document clustering, summarization, and
indexing.  Representations enabling accurate measurement of semantic
similarities between documents would greatly facilitate such
technologies.  In this paper, we focus on representations in which
vector directionality is used to represent a document's semantics, by
which we mean its (human-interpretable) constituent concepts.  This
goal is to be accomplished without access to concept labels, since
they are typically not available in many applications.

The vector space model (\vsm) is a classic method for constructing
such vector-based representations. It encodes a document collection by
a {\em term-document matrix} whose $[i,j]$th element indicates the
association between the $i$th term and $j$th document.  However, as
has been pointed out previously, \vsm\ does not always represent
semantic relatedness well; for instance, documents that do not share
terms are mapped to orthogonal vectors even if they are clearly
related.

{\em Latent Semantic Indexing (LSI)}
\cite{Deerwester+al:90a,Dumais:91a} attempts to overcome this and
other shortcomings by computing an approximation to the original
term-document matrix; this is equivalent to {\em projecting} the
term-document matrix onto a lower-dimensional subspace. LSI has been
successfully applied to information retrieval and many language
analysis tasks (e.g. \newcite{Dumais+Nielsen:92a},
\newcite{Foltz+Dumais:92a}, \newcite{Berry:95a},
\newcite{Foltz+Kintsh+Landauer:98a}, and \newcite{Wolfe+al:98a}),
prompting several studies to explain its effectiveness
(e.g. \newcite{Bartell+al:92a}, \newcite{Story:96a},
\newcite{Ding:99a}, \newcite{Papadimitriou+al:00a}, and
\newcite{Azar+al:01a}).

\newcite{Ando:00L} introduced an alternative subspace-projection
method, which we call {\em Iterative Residual Rescaling} (\ours), that
outperforms \lsi\ by counteracting its tendency to ignore
minority-class documents.  This is done by repeatedly rescaling
vectors to amplify the presence of documents poorly represented in
previous iterations.  However, Ando presented only heuristic arguments
to explain \ours's success.

%\smallskip

\paragraph{Contributions}  In this paper, we use the notion of subspace projection to formalize
the document representation problem.
Based on this framework, we provide a new theoretical analysis that
shows a precise relationship between the performance of LSI and the
{\em uniformity} of the underlying distribution of documents over
topics.  As a consequence, we provide an explanation for \ours's
success: {the rescaling it performs compensates for non-uniformities
in the topic-document distribution}.  Moreover, our framework yields a
new way to automatically adjust the amount of rescaling by estimating
the non-uniformity.

To support our theoretical results, we present performance
measurements both on document sets in which the topic-document
distributions were carefully controlled, and on unrestricted datasets
as would be found in application settings.  In all cases and for all
metrics, the results 
confirm our theoretical predictions.  For instance, \ours\ combined
with our new parameter selection technique achieved up to 10.1\%
higher {\em kappa average precision} than LSI and up to 8.7\% better
document clustering performance.  The experiments as a whole provide
strong evidence for the usefulness of our framework in general and
the effectiveness of our augmented \ours\ in particular.

\paragraph{Notational conventions} ~ A bold uppercase letter
(e.g. $\mv{M}$) denotes a matrix; the corresponding bold lowercase
letter with subscript $i$ (e.g.  $\mv{m}_i$) denotes the matrix's
$i$th column vector.  We use $\Range{\mv{M}}$ to denote $\mv{M}$'s
range, or column space: $\qset{\mv{y}}{\exists \mv{x} \mbox{ such that
}\mv{y} = \mv{M} \mv{x}}$. When a document collection has been
specified, $\numdocs$ denotes the number of documents the collection
contains.

%%%%%%%%%%%%%%%%%%%%%%
\section{Analyzing LSI}
\subsection{Topic-based similarities}

Our framework for analyzing Latent Semantic Indexing and other
subspace-based methods revolves around the notion of topic-based
similarity.  Fix an $\numdocs$-document collection $\collection$ and
corresponding $m$-by-$\numdocs$ term-document matrix $\mAhat$.  We
assume that there exists a set, denoted $\colltopics$, of
$\numtopics<\numdocs$ topics underlying $\collection$. We also assume
that for each topic $t$ and document $d$ there exists a (real-valued)
{\em relevance score} $\rel{t}{d}$, suitably normalized so that for
each $d$, we have $\sum_{t \in \colltopics} {\rel{t}{d}}^2 = 1$.  We
then define the {\em true topic-based similarity} between two
documents $d$ and $d'$ as:
$${\rm sim}(d,d') = \sum_{t \in \colltopics} \rel{t}{d}
\rel{t}{d'}\,.$$ 
It is convenient to summarize these similarities in
a single $\numdocs$-by-$\numdocs$ matrix $\simmatrix$, where
$\simmatrix[d,d'] = {\rm sim}(d,d')$.

Note that although we assume the existence of underlying topics as the
basis for the true document similarities, in contrast to other
analyses (e.g. \newcite{Papadimitriou+al:00a}, \newcite{Azar+al:01a},
and \newcite{Ding:99a}) we do
{\em not} assume that there is an underlying generative or
probabilistic model that {\em creates} the term-document matrix
$\mAhat$.

\subsection{The optimum subspace}
\label{sec:why}

We formulate the ultimate goal of subspace-based algorithms, such as
LSI, as choosing some subspace such that projecting $\mAhat$ onto this
subspace creates new document vectors whose measured similarities
(i.e., cosines) more closely correspond to the true topic-based
similarities.

More formally, for any subspace $\spaceAny \subseteq \Re^m$, the
unique {\em (orthogonal) projection} of a vector $\mv{x} \in \Re^m$
onto $\spaceAny$ is given by $\oProj{\spaceAny}(\mv{x}) = \mBase
\mBase^T \mv{x}$ for any $\mBase$ whose columns form an orthonormal
basis for $\spaceAny$.  We define the projection of $\mAhat$ onto
$\spaceAny$ as the matrix 
$\makematrix{\oProj{\spaceAny}(\vAhat_1)\cdots\oProj{\spaceAny}(\vAhat_n)}$, 
i.e., the result of projecting
each of the term-document vectors.  Hence, after projection onto
$\spaceAny$, the similarity between the $i$th and $j$th documents is
measured by
$$\cos(\oProj{\spaceAny}(\vAhat_i),\oProj{\spaceAny}(\vAhat_j)) =
\frac{\oProj{\spaceAny}(\vAhat_i)^T\oProj{\spaceAny}(\vAhat_j)}
{|\oProj{\spaceAny}(\vAhat_i)| |\oProj{\spaceAny}(\vAhat_j)|}\,.$$
The {\em document representation problem} is as follows:  given
$\mAhat$ --- but {\em not} $\simmatrix$ or even any knowledge of what
the underlying topics are --- find a subspace
$\spaceAny$ such that the entries of the {\em deviation
matrix}\footnote{It suffices to consider the inner products
$\oProj{\spaceAny}(\vAhat_i)^T\oProj{\spaceAny}(\vAhat_j)$ rather than the
cosines because, as shown by \newcite{Ando:thesis}, if there exists
$\epsilon <1$ that upper-bounds the magnitudes of the deviation
matrix's entries, then for any  $\vAhat_i$ and
$\vAhat_j$,
$$ \frac{{\rm sim}(i,j) - \epsilon}{1 + \epsilon} \le
\cos(\oProj{\spaceAny}(\vAhat_i), \oProj{\spaceAny}(\vAhat_j)) \le
\frac{{\rm sim}(i,j) + \epsilon}{1 - \epsilon} \,.$$
} 
$$\fError(\spaceAny) = \simmatrix - \oProj{\spaceAny}(\mAhat)^T
\oProj{\spaceAny}(\mAhat)$$ are small.  The {\em optimum
subspace}
$$\spaceOpt = \mathop{\arg\min}_{\spaceAny \subseteq \Range{\mAhat}}
\nTwo{\fError(\spaceAny)}\,,$$ with ties broken by smallest
dimensionality and then arbitrarily, serves as the standard for
comparison in our analysis.  We denote the corresponding projection
operator by $\oProj{opt}$, and use $\sOptErr$ to denote the {\em
optimum error} $\nTwo{\fError(\spaceOpt)}$.  Note that $\sOptErr$ need
not be zero, as it may be impossible to project the given
term-document matrix in such a way as to perfectly recover the true
document similarities.

\subsection{The singular value decomposition  and \lsi}
\label{sec:svd}

In this section, we first  briefly introduce the {\em singular value
decomposition} (SVD) \cite{Golub+VanLoan:96a}, since singular values 
are necessary for our analysis. Then, we describe LSI, which is based
on the SVD.

The SVD factors an arbitrary rank-$h$ matrix $\mv{Z} \in \Re^{r \times
s}$ into the following product:
$$ \mv{Z} = \mv{U} \mv{\Sigma} \mv{V}^T\,,$$
where the columns of $\mv{U} \in \Re^{r\times h}$ and
$\mv{V} \in \Re^{s \times h}$ are orthonormal,  $\mv{\Sigma} =
\mbox{diag}(\sigma_1, \sigma_2, \ldots, \sigma_h)$ is diagonal
(following convention, we assume $\sigma_1 \ge \cdots \ge \sigma_h$),
and the $\sigma_i$'s are all positive.
The quantities $\sigma_i$, $\mv{u}_i$ and $\mv{v}_i$ are called the
$i$th {\em singular value}, {\em left singular vector}, and {\em right
singular vector}, respectively.  The left singular vectors span
$\mv{Z}$'s range, and $\sigma_1 = \nTwo{\mv{Z}}$.

Zeroing out all but the $\ell < h$ largest singular values yields the
least-squares optimal rank-$\ell$ approximation to $\mv{Z}$.  {\em Latent
Semantic Indexing} (LSI) \cite{Deerwester+al:90a,Dumais:91a} applies
this rank-$\ell$ approximation to the term-document matrix $\mAhat$, which
corresponds to projecting $\mAhat$ onto the {\em rank-$\ell$ \lsi\
subspace} spanned by $\mv{u}_1, \ldots, \mv{u}_\ell$: letting
$\mv{U}_\ell = \makematrix{\mv{u}_1\ldots \mv{u}_\ell}$,
$$
\mv{U}_\ell \mv{U}_\ell^T  \mAhat = \mv{U} \;
        \mbox{diag}(\underbrace{\sigma_1,  \cdots, \sigma_\ell, 0,
        \cdots, 0}_{h \mbox{ entries}}) \;
        \mv{V}^T.
$$
Note that the fact that this matrix approximates
$\mAhat$ well does not imply that it represents the true document
similarities well, as we shall see.

Further intuition may be gained on the left singular vectors by the
following observation.  Let $\projj{j}(\vAhat_i)$ be the projection of
$\vAhat_i$ onto the span of $\mv{u}_1, \ldots, \mv{u}_{j}$, and let
$\residj{i}{j}$ be the {\em residual vector} $\vAhat_i -
\projj{j-1}(\vAhat_i)$.  Then, $\mv{u}_j$ is the unit vector that
maximizes the following quantity:
$$
\svdopt^{(j)}(\mv{x}) =
 \sum_{i=1}^n \left( |\residj{i}{j}| \cos(\residj{i}{j}, \mv{x}) \right)^2\,.
$$
In a sense, $\mv{u}_j$ resembles a weighted average of residual
vectors, where longer residuals receive greater
weight.  Hence, the $\ell$ left singular vectors may be thought of as
representing the $\ell$ major directions
in the document collection.

\subsection{Non-uniformity and LSI}
\label{sec:theorems}

We now state our results relating the non-uniformity of the underlying
topic-document distribution to the quality of the document
representation spaces derived using LSI.  The main outline of our
argument is to first show relations between certain singular values
and certain quantities linked to our topic model, and then show how
the distance between the LSI-subspace and the optimal subspace relates
to these singular values.  Proofs of these results, which make use of
invariant subspace perturbation theorems
\cite{Davis+Kahan:70a,Stewart+Sun:90a,Golub+VanLoan:96a}, are
sketched in the appendix of this paper and given in full in
\newcite{Ando:thesis}.

Recall that we are dealing with a fixed document collection
$\collection$ with $k$ underlying topics.  Throughout, we use $h$ to
denote the dimensionality of the optimum subspace.  For clarity, we
will abuse notation by writing ``$x \in y \pm z$'' as shorthand for
``$x \in [y-z,y+z]$''.

A crucial quantity in our analysis is the {\em dominance} $\pop{t}$ of
a given topic $t$: $$\pop{t}= \sqrt{\sum_{d \in \collection}
\rel{t}{d}^2}\,.$$ (It may be helpful to observe that in the special
``single-topic documents'' case, where each document is relevant to
only one topic in $\colltopics$, squaring $\pop{t}$ gives exactly the
number of documents in $\collection$ that are relevant to topic $t$.)  We
assume without loss of generality that $\pop{1} \ge \pop{2} \ge \cdots
\ge \pop{k}$, and for convenience set $\pop{i}=0$ if $i>k$.

Now we show that the projection of $\mAhat$ onto the optimum subspace
$\spaceOpt$ in some sense reveals the topic dominances.  It is
intuitively clear, however, that the extent to which this holds should
depend to some degree both on the optimum error $\sOptErr$ and on the
{\em topic mingling} $\sTsim = \left(\sum_{t,t' \in \colltopics, t \ne
t'} \left(\sum_{d \in \collection} \rel{t}{d}
\rel{t'}{d}\right)^2\right)^{1/2}$ (note that in the single-topic
documents case, $\sTsim = 0$).  Certainly, if the optimum error is
high, then we cannot expect the optimum subspace to fully reveal the
topic dominances; also, if there is high topic mingling in the
collection, then the topics will be fairly difficult to distinguish.
\begin{theorem}
\label{theorem_population}
Let $\sv_i$ be the $i$th largest singular value of
$\oProj{opt}(\mAhat)$.  Then, $\sv_i^2 \in \pop{i}^2 \pm (\sOptErr + \sTsim)$.
\end{theorem}
This result gives us leave to define $\svMax = \sv_1$ and $\svMin =
\sv_h$, where $h$ is the dimension of $\spaceOpt$.  The ratio
$\svMax/\svMin$ then serves as a measure of the {\em
non-uniformity} of the topic-document distribution underlying the
collection:  
the more the largest topic dominates the collection,
the higher this ratio will tend to be.

Now we are in a position to present our main theorem. This result
bounds the distance between the optimum subspace $\spaceOpt$ and the
same-dimensionality LSI subspace $\spaceLSI$ by a function of the
non-uniformity of $\collection$'s topic-document distribution.  The
bound also incorporates a certain value (defined precisely in the
appendix) $\sOrgErrapprox \in \sOrgErr \pm \sOptErr$, where 
$\sOrgErr= \nTwo{\fError(\spaceOrg)}= \nTwo{\simmatrix - \mAhat^T \mAhat}$ is the {\em input error}:
intuitively, if a ``bad'' term-document matrix is received as input,
one cannot expect LSI to do well.
\begin{theorem}
\label{theorem_1}
Let $\spaceLSI$ be the $h$-dimensional LSI subspace spanned by the the
first $h$ left singular vectors of $\mAhat$.  If $\svMin >
\sqrt{\sOrgErrapprox}$, then
$$
\nTwo{\tan(\Theta(\spaceLSI,\spaceOpt))} \leq \frac{\svMax}{\svMin} \cdot
\frac{\sqrt{\sOrgErrapprox}/\svMin}{1 -
(\sqrt{\sOrgErrapprox}/\svMin)^2}\,,
\label{eqn:tan_bound}
$$
\hspace{-.04in}
where $\Theta$ is the {\em canonical angle matrix} measuring the distance
between subspaces \cite{Davis+Kahan:70a,Stewart+Sun:90a}.
\end{theorem}

Intuitively, what this means is that {\em $\spaceLSI$ must be close to
$\spaceOpt$ when the topic-document distribution is relatively
uniform} and the input error is small in comparison to the $h$th
largest topic dominance.  Conversely, if the input error is fixed,
{\em our bound on LSI's performance weakens when the underlying
topic-document distribution is highly non-uniform}. Finally, we note
that the condition on $\sOrgErrapprox$ is natural: roughly
speaking, if the input error is large enough to ``swamp'' the
dominance of the $h$th largest topic, then intuitively we cannot
expect good results.

Finally, we note that a related result can be proved which, roughly
speaking, links lower bounds on the distance between the two
subspaces to non-uniformity and the input error; however, this theorem
is quite technical in nature and thus is omitted.
We refer the reader to \newcite{Ando:thesis} for details.

%*********************
\subsection{Related work: theoretical analyses of LSI}
\label{sec:comparison}

As noted above, there have been several studies analyzing \lsi, using
approaches such as Bayesian regression models \cite{Story:96a} and
Gaussian models \cite{Ding:99a}.  Here, we concentrate on describing
the work most similar in spirit to ours.

\newcite{Zha+Marques+Simon:98a} propose a subspace-based model
for LSI.  Their work focuses on dimensionality selection and
implementation issues regarding more accurate updating schemes.
\newcite{Bartell+al:92a} show that \lsi\ can be regarded
as a solution to the special Multidimensional Scaling (MDS) problem of
preserving the inner products of the original document vectors.  However, as
noted above, this is not the same as recovering hidden topic-based
similarities, especially in the case of noisy data.

Perhaps the work most similar to ours is that of 
\newcite{Papadimitriou+al:00a} and  \newcite{Azar+al:01a}, both
of which propose to explain LSI's success with analyses that, like
ours, employ invariant subspace perturbation theorems. Papadimitriou
et al. start with a probabilistic corpus model.  By assuming low input
error and certain conditions on singular values (which, from our
perspective, can be considered to be roughly equivalent to assuming
relative uniformity, although Papadimitriou et al. did not
explicitly make this connection), they show that LSI will work well
with high probability.  But their results are based on a {\em pure}
probabilistic corpus model in which all the documents are single-topic
and topics have associated {\em primary} (distinguishing) disjoint
sets of terms.  Thus, their analysis holds only for a very restricted
class of document collections.  Similarly, Azar et al.  also start
with particular conditions and a specialized underlying generative
model to show that LSI works well for ``good'' documents with high
probability.  In contrast to these approaches, our analysis does not
assume a model of term-document matrix creation, and so applies to
{\em arbitrary} term-document matrices, with the non-uniformity and
the input matrix's quality being explicit terms in our bound.

\section{\hspace*{-.1in}\ours:  Overcoming non-uniformity}
\label{sec:method}

Our results from Section 
\ref{sec:theorems} 
indicate that we could improve the
performance of \lsi\ if we could somehow ``smooth'' the topic-document
distribution (that is, effectively lower
$\svMax/\svMin$).  
We show that the {\em Iterative Residual Rescaling} (\ours) algorithm,
introduced (but not named) and heuristically motivated by
\newcite{Ando:00L}, accomplishes this task {\em without prior
knowledge of the assignments of documents  to topics}.
\subsection{Ando's IRR algorithm}

\begin{figure*}[th]
\epsfscaledbox{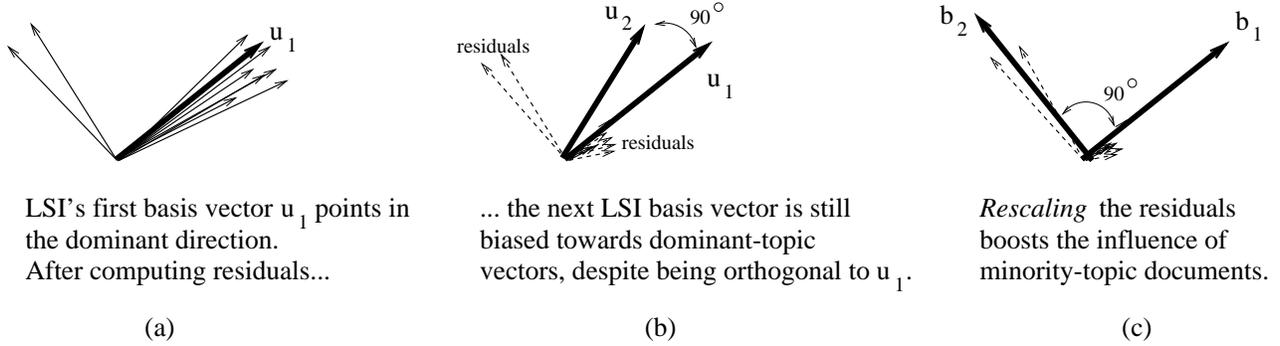}{6.7in}
\caption{\label{fig:compensation}
Effect of nonuniformity on \Lsi, and how IRR compensates.}
\end{figure*}

\begin{figure*}[htb]
\begin{center}
\begin{tabbing}
\hspace*{1in} \={\bf \ours}\=$(q,$\=$\irrdim)$:\hspace*{.05in} \= \hspace*{.05in} \=\\
\> \> $\mv{R}$:= $\mAhat$ ~~~ /* initialize residuals by
given $m$-by-$\numdocs$ term-document matrix */ \\
\> \> For $j := 1, 2, \ldots, \irrdim ~~~~$ /* create $\irrdim$ basis vectors */\\
\>\> \> For $i:=1, 2, \ldots, \numdocs$ \\
\> \> \> \> ${\hat{\mv{r}}_i} := |\mv{r}_i|^q \mv{r}_i ~~~~$ /* rescale
residuals */\\
\> \>  \> $\mv{b}_j := \arg\max_{\mv{x}: |\mv{x}| = 1} \left(\sum_{i=1}^\numdocs
\left( |{\hat{\mv{r}}_i}| \cos(\hat{\mv{r}}_i, \mv{x}) \right)^2 \right)$ \\
\>\> \> For $i:=1, 2, \ldots, \numdocs$ \\
\>\> \> \> Subtract from $\mv{r}_i$ its projection  onto $\mv{b}_j$
~~~ /* recompute residuals */\\
\> \>  $\mAhat_{\ours}:= \mv{B}\mv{B}^T \mAhat$, where  $\mv{B} =
\makematrix{\mv{b}_1 & \mv{b}_2 & \cdots & \mv{b}_{\irrdim}}$ /* new document representation */
\end{tabbing}
\end{center}
\caption{\label{fig:pseudo_code} High-level pseudocode for Ando's
\ours\ (i.e. before augmentation with \asfd).}
\end{figure*}

Recall from Section \ref{sec:svd} that the left singular vectors
$\mv{u}_1,$ $\mv{u}_2, \ldots$ produced by
LSI can be derived, one by one, via the following computation:
$$
\mv{u}_j = \mathop{\arg\max}_{|\mv{x}|=1} \sum_{i=1}^\numdocs 
\left( |\residj{i}{j}| \cos(\residj{i}{j}, \mv{x}) \right)^2,
$$
where the $\residj{i}{j}$ are the residuals $\mv{d}_i -
\projj{j-1}(\mv{d}_i)$.  Unfortunately, inspection of this formula
shows that when the topic-document distribution is highly non-uniform,
the cumulative influence of a large number of (small) residuals for a
major topic can cause smaller topics to be ignored, as depicted in
Figures \ref{fig:compensation}(a) and (b).

Our explanation of \ours's effectiveness is that by amplifying the
length differences among residual vectors, \ours\ boosts the influence
of minority-topic documents, thus {\em compensating for non-uniform
topic distributions} (Figure \ref{fig:compensation}(c)).  More
precisely, \ours, like (our formulation of) the SVD, computes basis
vectors by successive maximizations, as shown in the pseudocode in
Figure \ref{fig:pseudo_code}.  Crucially, though, \ours's objective
function, $\oursopt^{(j)}$, incorporates a scaling factor $q$ via the
scaling function $\irrscale{\mv{r}}{q} = |\mv{r}|^q~\mv{r}$:
$$
\oursopt^{(j)}(\mv{x}) = \sum_{i=1}^\numdocs 
\left( |\irrscale{\residj{i}{j}}{q}|~\cos(\irrscale{\residj{i}{j}}{q}, \mv{x}) \right)^2\,.
$$
This is maximized by the first left singular vector of
$$
\widehat{\mv{R}}^{(j)} =
\makematrix{\irrscale{\residj{1}{j}}{q} & \cdots &
\irrscale{\residj{n}{j}}{q} }\,.  
$$
That is, \ours\ {\em rescales} each residual vector $\mv{r}_i$ at each
basis vector computation, increasing the contrast between long and
short residuals when $q > 0$.  LSI is the special case in which $q=0$.

\subsection{The AUTO-SCALE method}
\label{sec:scaling_factor}

Our discussion above argues that the degree of rescaling should
depend on the uniformity of the topic-document distribution.  
\newcite{Ando:00L} did not explicitly make this connection, and hence
could not take advantage of it: $q$ was determined simply by training
on held-out data.  In contrast, our novel analysis allows us to
exploit this connection to develop an effective estimation method ---
{\em automatic scaling factor determination} (\asfd) --- that
approximates the topic-document non-uniformity without
prior knowledge of the underlying topics.

\asfd\ is based on the observation that we can use the quantity
$\sum_{t \in \colltopics} \pop{t}^4 / \numdocs^2$ as a measure of the
non-uniformity of the topic-document distribution.  Of course, we
don't have access to the topic dominances $\pop{t}$, but we can
approximate this measure by
\footnote{The {\em Frobenius norm} $\normF{\mv{X}}$ is defined as
$\sqrt{\sum_{i,j}\mv{X}[i,j]^2}$.}
$$
\nonuniapprox(\mAhat) \stackrel{def}{=} \left(\frac{||\mAhat^T \mAhat||_F}{\numdocs} \right)^2 \,.
$$
This approximation follows from first assuming that 
the input matrix is fairly good, so that
$||\mAhat^T \mAhat||_F^2$ is roughly
equal to $||\simmatrix||_F^2 = \sum_{d,d' \in \collection }
\left(\sum_{t \in \colltopics} \rel{t}{d} \rel{t}{d'} \right)^2$.
The latter can be rewritten, after some algebra, as the quantity
$\sum_{t \in \colltopics} \pop{t}^4 + \sTsim^4$, which is roughly
$\sum_{t \in \colltopics} \pop{t}^4$ if we assume approximately
single-topic documents.  In practice, we set $q$ to a linear function
of $\nonuniapprox(\mAhat)$; this is discussed in Section
\ref{sec:controlled_data}.

Although the above assumptions  are rather coarse,
\asfd\ yields good empirical results: see  Sections
\ref{sec:controlled} and \ref{sec:evaluation}.

\subsection{Dimensionality selection}
\label{sec:dimensionality}

\ours's second parameter is $\irrdim$, the dimensionality of the
created subspace.  One way to set this parameter is to train it on
held-out data.  Following \newcite{Ando:00L}, we found that learning
thresholds on the {\em residual ratio} $||\mv{R}^{(j)}||_F^2 /
\numdocs$ as a stopping criterion is effective for both \lsi\ and
\ours.  Intuitively, this ratio describes how much is left out of the
proposed subspace (of course, we do not want to reproduce the
term-document matrix exactly --- hence the threshold).  Note that this
training method allows some flexibility in the chosen dimensionality:
for different data sets, the same residual ratio threshold may result
in selecting a different $\ell$.

While training on held-out data is reasonable and is commonly employed
in practice, it is a relatively expensive process.
A speedier alternative arises in settings in which $k$, the
number of topics, is pre-specified --- examples include cases where the
topic set is a fixed class such as the TREC topic labels,
or where the application allows the user to specify the appropriate
level of granularity for his or her needs.   In such settings, we could
simply set the dimensionality equal to $k$ as a matter of
convenience. (Indeed, \newcite{Papadimitriou+al:00a} show that under
certain strong assumptions on the data, rank-$k$ \lsi\ should perform
well.  See also \newcite{Ando:thesis}.)
We describe experiments with both selection methods below.

\section{Evaluation metrics}
\label{sec:framework}

\noindent {\bf Kappa average precision} ~ Our first evaluation metric
is based on the {\em pair-wise average precision} \cite{Ando:00L},
adapted from the average precision measure commonly used in
information retrieval. The motivation behind this metric is that the
measured similarity for any two {\em intra-topic} documents (i.e.,
that share at least one topic) should be higher than for any two {\em
cross-topic} documents which have no topics in common.  More formally,
let $p_i$ denote the document pair with the $i$th largest measured
similarity (cosine).  Precision for an {\em intra-topic} pair $p_j$ is
defined by
$$\precn(p_j) = \frac{\# \mbox{ of intra-topic pairs $p_i$ such that $i
\le j$}}{\mbox{$j$}}\,.$$
The {\em pair-wise average precision} is
the average of these precision values over all intra-topic pairs.

To compensate for the effect of large topics (which increase the
likelihood of chance intra-topic pairs), we modify the pair-wise
average precision to create a new metric, which we call the {\em kappa
precision} in reference to the Kappa statistic
\cite{Siegel+Castellan:88a,Carletta:96a}:
$$
\precn_\kappa(p_i) = \frac{\precn(p_i) - {\rm chance}}{1 -
{\rm chance}} \,,
$$
where ${\rm chance} =$ ($\#$ of intra-topic pairs)/($\#$ of document pairs).
The {\em kappa average precision} $\kappa$ is defined to be the average of the
kappa precision over all intra-topic pairs, and is a linear function
of the pair-wise average precision.

\smallskip
\noindent {\bf {Clustering}} ~ We also test how well the new subspaces
represent document similarities by seeing whether document clustering
improves when these new representations are used as input.
To simplify the scoring, we consider only single-topic documents.

Let $\conting$ be a cluster-topic contingency table such that
$\conting[i,j]$ is the number of documents in cluster $i$ that are
relevant to topic $j$, as in \newcite{Slonim+Tishby:00a}.  We define
$\revalx(\conting) = \sum_{i,j} \assoc_{ij}/n$, where 
$\assoc_{ij} =
\conting[i,j]$ if $\conting[i,j]$ is the {\em unique} maximum in both
its row and column, and $\assoc_{ij} = 0$ otherwise.
Note that this (rather strict) measure only considers the most tightly
coupled topic-cluster assignment, and decreases when either cluster
purity or topic integrity 
declines (see Figure
\ref{fig:contingency_table}).

\begin{figure}[t]
\begin{center}
\begin{tabular}{|l|l|l|l|l|}
\hline
          & topic 1 & topic 2 & topic 3 & topic 4 \\
\hline
cluster 1 &     5   &    10   & \bf{20} &     0   \\
\hline
cluster 2 &     5   &    10   &     5   &     0   \\
\hline
cluster 3 &     0   &     0   &     0   & \bf{21} \\

\hline
cluster 4 & \bf{15} &     5   &     0   &     0   \\
\hline
cluster 5 &     0   &     0   &     0   &     4   \\
\hline
\end{tabular}
\end{center}
\caption{\label{fig:contingency_table} Sample contingency table, with
 $\revalx(\conting) = (15 + 20 + 21) / 100 = 56 \%$.
}
\end{figure}

To factor out the idiosyncracies of particular clustering algorithms,
we apply six standard clustering methods
--- single-link, complete-link, group average, and k-means with
initial clusters generated by these three methods ---
to the document vectors in each proposed subspace,
and record both the {\em \high} (highest) and {\em \low} (lowest)
$\revalx(\conting)$ scores.
While the \high\ performance is perhaps more intuitive,  we observe
that the \low\
performance also gives us important information about 
the quality of the representation being evaluated: if the \low\ is
low, then there is at least one clustering algorithm for which the
document subspace is not a good representation; otherwise, the
representation is good for {\em all six} clustering algorithms.

\section{Controlled distributions}
\label{sec:controlled}

Our first suite of experiments studies the dependence of \lsi\ and
\ours\ on increasingly less uniform topic-document distributions.  The
results strongly support our theoretical analysis of \lsi's
sensitivity to non-uniformity.

\subsection{Experimental setting}
\label{sec:controlled_data}

To focus on distributional non-uniformity, we first chose two TREC topics,
and then specified seven {\em distribution types}: (25, 25), (30, 20),
(35, 15), (40, 10), (43, 7), (45, 5), and (46, 4), where $(n_1,n_2)$
indicates that $n_i$ of the documents are relevant to topic $i$.  For
each of these types, we generated ten sets of 50 TREC documents each,
where each document was relevant to exactly one of the pre-selected topics.
We also created five-topic\footnote{Results for three- and four-topic
document sets were similar and are therefore omitted.}  data sets in the same
manner, using distribution types of the form $(i,j,j,j,j)$ (which makes
uniformity comparisons obvious).

To create the term-document matrices, we extracted single-word stemmed
terms using {\sc TALENT} \cite{Boguraev+Neff:00a}, removed stopwords, 
and then length-normalized the document vectors (so that term weights
were frequency-based).

To implement \asfd, we set $q=\alpha \cdot \nonuniapprox(\mAhat) +
\beta$, where $\alpha=3.5$ and $\beta=0$ for {\em all} our
experiments.  These values (which are necessary to determine the
``units'' of the scale factor) were empirically determined once and
for all from observations on data disjoint from our test sets.  This
contrasts with training $q$ for every new test set encountered, as in
\newcite{Ando:00L}.  Training is an expensive process, and we envision
interactive applications such as organizing query results (a task we
simulate in Section \ref{sec:evaluation}) in which what would serve as
training data is not obvious.  We thus view \asfd\ as a practical
alternative to the usual parameter training.

For simplicity, the dimensionality of \lsi\ and \ours\ in our
experiments was set to the number of topics.\footnote{
Our preliminary experiments with dimensionality training
indicated that indeed this was often the best
dimensionality for \lsi\ and almost always the best for \asfd-\ours.}

\subsection{Controlled-distribution results}
\label{sec:controlled_eval}

\begin{figure}[t]
\epsfscaledbox{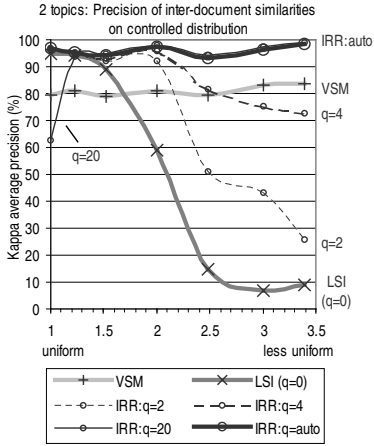}
{2in}
\caption{\label{fig:c2_ak} Kappa average performance, two topics. Points are averages over ten document sets.}
\end{figure}
\begin{figure}[t]
\epsfscaledbox{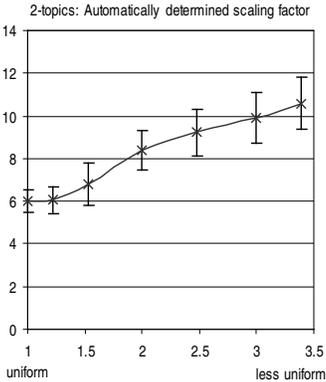}{1.8in}
\caption{\label{fig:c2_sf}  Automatically-determined scaling factor
values: ten-set average and standard deviation.}
\end{figure}

We first examine the kappa average precision results, shown in Figure
\ref{fig:c2_ak}.    The $x$-axis represents the nonuniformity of the
topic-document distribution, as measured by $\popMax/\popMin$.
We see that when the topic-document distribution is relatively
uniform, \lsi's performance is higher than 90\%.  However, as the
nonuniformity increases, the performance of \lsi\ drops precipitously,
in accordance with our theorems above.

Also, our interpretation of the scaling factor $q$ as
compensating for non-uniformity is borne out nicely. For highly uniform
distributions, the performance difference between $q=0$ (at which
\ours\ = \lsi), $q=2$, and $q=4$ is not great.  At medium
nonuniformity, $q=0$ degrades, but $q=2$ still does about the same as
$q=4$.  But as the non-uniformity increases even more, we see that
$q=2$ is not large enough to compensate, and so declines in comparison
to $q=4$.  

Furthermore, we see that \ours\ with \asfd\ (labelled
`{\ours}:$q={\rm auto}$') does extremely well across all levels of
non-uniformity.  Figure \ref{fig:c2_sf} shows that \asfd\ 
indeed adjusts for more non-uniform distributions: the chosen
scaling factor increases on average as the non-uniformity goes up.

Now, one might conjecture  that instead of using
\asfd, it would suffice simply to choose a single very large value of
$q$.  Intuitively, though, this is problematic, since too high a
scaling factor would tend to
completely eliminate  residuals.  Furthermore, the $q=20$ curve in
Figure \ref{fig:c2_ak} disproves the conjecture: 
in the uniform case, selecting an overly large scaling factor hurts
performance, driving it below the baseline \vsm\ curve.

\begin{figure}\hspace*{-.2in}
\epsfscaledbox{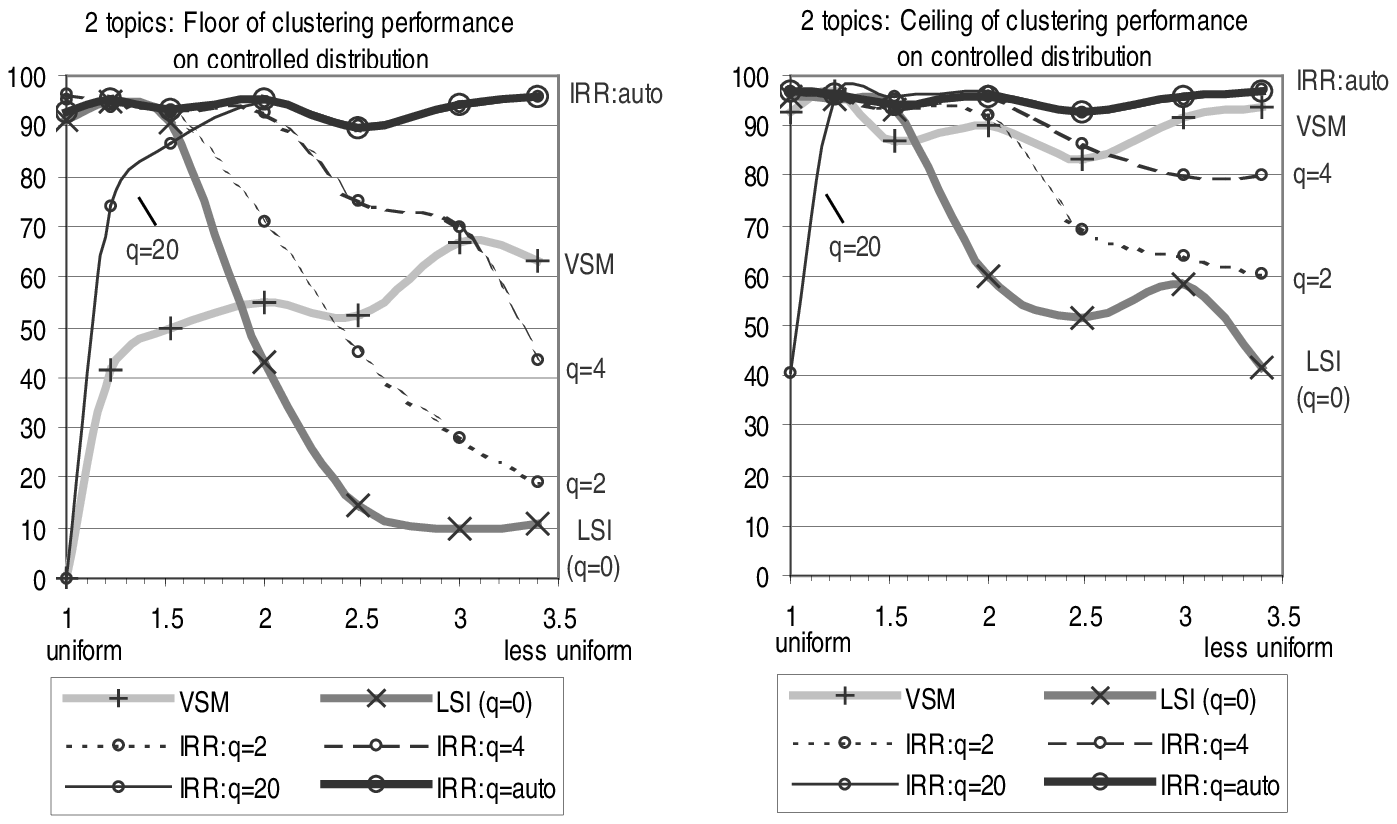}
{4in}
\caption{\label{fig:c2_cluster} Floor and ceiling clustering results,
two topics. Points are averages over ten document sets.}
\end{figure}

The two-topic floor and ceiling clustering results, shown in Figure
\ref{fig:c2_cluster}, exhibit precisely the same types of behaviors as
in the kappa average precision case.  The floor performances are
especially interesting, as they show that \asfd-\ours\ exhibits very
good performance for all six of our rather wide variety of clustering
algorithms.  They also indicate that \vsm\ is `fragile' for uniform
distributions, in that sometimes it is a very poor representation for
at least one of the clustering algorithms we employed.

\begin{figure*}[ht]
\epsfscaledbox{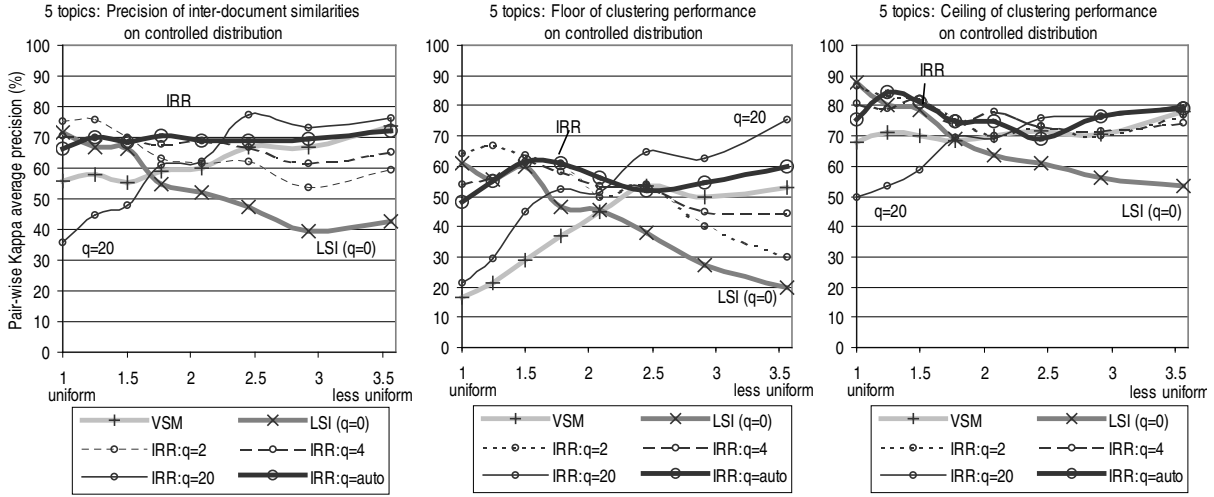}
{6.4in}
\caption{\label{fig:c5} Ten-set averages of kappa average precision
and floor and ceiling clustering results, five topics.}
\end{figure*}

Finally, Figure \ref{fig:c5} shows the results of the same evaluation
experiments run on five-topic data.  Again, the empirical results
are completely in line with what we predicted, with \asfd\ leading to
strong performance over all metrics and all degrees of
non-uniformity.  Note that the gap between \lsi\ and \vsm\ decreases
in comparison to the $k=2$ case; this is due to the fact that at
higher dimensionalities, the subspace produced by \lsi\ gets closer to
that of the original term-document matrix.

These results all strongly support our theoretical claims.

\vspace*{-.1in}

\section{Unrestricted distributions}
\label{sec:evaluation}

In this section, we experiment on the more realistic setting of
document sets without distribution restrictions. We expect that in
practice, topic-document distributions will be fairly
non-uniform, so that \ours\ should perform well in comparison to \lsi.
Figure \ref{fig:eval_summary} summarizes the evaluation settings.

We used 648 TREC documents, each relevant to exactly one of twenty
TREC topics.  To perform parameter training, we randomly divided these
documents into two disjoint document pools.  We then simulated input
from an information retrieval application by generating 15 document
sets from each pool, where each set consisted of those documents
containing one of 15 arbitrarily chosen keywords; this yielded a total
of 30 document sets.  Document sets from one pool were used as
parameter training data for the sets from the other pool, and vice
versa.  Performance results are averages over these 30 runs.  The
scaling factor for \ours\ was determined by \asfd\ in all cases (again
with the same constants $\alpha$ and $\beta$ as before).  
The term-document matrices were created in the same manner as in 
Section \ref{sec:controlled_data}.

\subsection{Kappa average precision results}
\label{sec:eval_ak}

Recall that we consider two ways to choose the dimensionality of a
document subspace.  In the first case, the system knows $k$, the
number of topics underlying the collection (in practice, this
information could be user-supplied as a way to control topic
granularity, or given by a set of predetermined classification
labels), and sets the dimensionality to it.  In the second case, $k$
is considered unknown, so we simply train the dimensionality parameter
using the residual ratio method described in Section
\ref{sec:dimensionality}.

\newcommand{\trained}{train}

\begin{figure}[t]
\begin{center}
\begin{tabular}{|c||c|c|c|c|c|}
\hline
Metric                  &  \multicolumn{2}{|c|}{$\kappa$}
                        &  \multicolumn{3}{|c|}{floor/ceiling}        \\
\hline
\hline
Is $k$ given?&  \multicolumn{1}{|c|}{yes}        &  no &
\multicolumn{2}{|c|}{yes}        & no  \\ \hline
Choice of $\ell$         &  $k$ &  \trained  & $k$   & \trained         & \trained \\
\hline
\# of clusters          &  \multicolumn{2}{|c|}{N/A} &
\multicolumn{2}{|c|}{$k$} &$\ell$ \\  \hline
Section (Data)        &  \multicolumn{2}{|c|}{\ref{sec:eval_ak} (Table \ref{fig:eval_ak})}  &  \multicolumn{3}{|c|}{\ref{sec:eval_cluster} (Figure \ref{fig:eval_cluster})} \\ \hline
\end{tabular}
\end{center}
\caption{\label{fig:eval_summary} Evaluation settings for unrestricted
distributions. Recall that $k$ is the number of topics and $\ell$ is
the dimensionality.}
\end{figure}

From Figure \ref{fig:eval_ak}, we see that
\ours\ yields higher $\kappa$ than \lsi\ and \vsm\ for
both dimensionality selection methods, and therefore does a better job
at representing inter-document similarities.
\lsi\ performs relatively poorly on this task; indeed, using $k$
dimensions in the \lsi\ case leads to worse results than \vsm.

\begin{figure}[bht]
\begin{center}
\begin{tabular}{|l||ll|} \hline
Dimensionality selection	& \lsi & \ours \\ \hline
Number of topics & -8.7 & 1.4 \\
Trained & 0 & 4.0 \\ \hline
\end{tabular}
\caption{\label{fig:eval_ak}
Thirty-set average absolute improvement in $\kappa$ over \vsm\ (51.4\%),
unrestricted distributions.}
\end{center}
\end{figure}

\subsection{Clustering results}
\label{sec:eval_cluster}

To derive \low\ and \high\ clustering performance results, there are
two parameters we need to specify: the dimensionality of the
subspace, and the  number of clusters.

\begin{figure}[t]
\epsfscaledbox{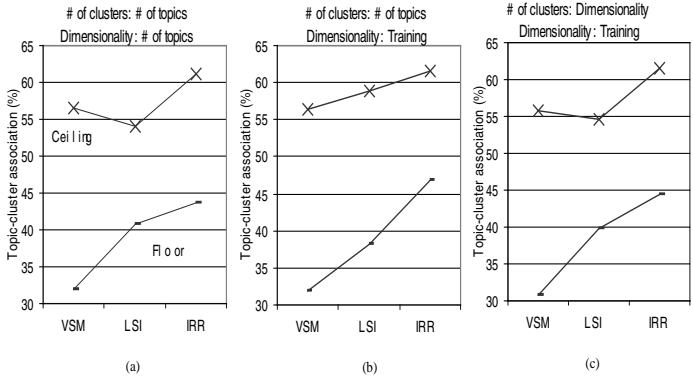}{3.7in}
\caption{\label{fig:eval_cluster}
Document clustering performances, unrestricted distributions:
averages over 30 runs.  }
\end{figure}

If $k$, the number of topics, is available, then it is the natural
choice for the number of clusters.  Then, to choose the dimensionality
in this case, one option is to also set it to $k$; Figure
\ref{fig:eval_cluster}(a) shows the results.  We see that IRR has the
best clustering performance overall.  Note that LSI's ceiling is
actually lower than VSM's.

When $k$ is given but we train the dimensionality via the residual
ratio, 
\ours\ still provides a better subspace for all the 
clustering algorithms we considered, both in terms of floor and
ceiling performance (Figure \ref{fig:eval_cluster}(b)).  We observe
that for this type of data, training the dimensionality allows \lsi\
to produce improved \high\ results.

We now consider the case in which $k$ is unknown. In this
situation, we know of no alternative but to train the dimensionality on
held-out data. As for the number of clusters, a reasonable default is
to simply set this value to the trained dimensionality.  Of course,
this doesn't apply to \vsm, since the dimensionality is not a free
parameter for it; instead, we set the number of clusters to the
average of the number of topics in the training document sets.

Figure \ref{fig:eval_cluster}(c) shows the clustering results for the
unknown-$k$ setting. \lsi's \high\ degrades by 4.3\% compared with
when the number of topics is given, while those of \vsm\ and \ours\
show almost no change.
Furthermore, \ours\ clearly outperforms the other methods.

\subsection{Discussion}
\label{sec:eval_discussion}

In our experiments, \lsi\ did worse or essentially the
same as \vsm\ in 4 out of 8 combinations of practical settings and
metrics.  In particular, when the dimensionality is chosen to be the
number of topics, \lsi\ performs relatively poorly.  Dimensionality
training improves \lsi's kappa average precision scores, and also
improves its clustering performance with respect to \vsm\ as long as
the correct number of clusters (i.e. the number of topics) is given.
However, when the number of clusters is unknown, \lsi's \high\
clustering performance drops, again indicating that for \lsi\, the
dimensionality should not be tied to the number of clusters.

In contrast, \ours\ consistently performs better than \lsi\ and \vsm\
for all our settings and metrics.  In particular, \ours\
fares relatively well when the dimensionality is set to the number of
topics as compared to when the dimensionality is actually trained.
These results suggest that setting the dimensionality to the number of
topics, when known, may be a practical alternative to dimensionality
training. Furthermore, in clustering applications for which the number
of topics is not known, we at least might be able to reduce the
training effort by only searching for the dimensionality, setting the
number of clusters to the same value.

\section{Conclusion}
\label{sec:conclusion}

To conclude, we review our three main results.  First, we have
provided a new theoretical analysis of \lsi, showing a precise
relationship between \lsi's performance and the uniformity of the
underlying topic-document distribution.  Second, we have used our
framework to extend Ando's (2000) \ours\ algorithm by giving a novel
and effective method for determining the requisite scaling factor.
Third, we have shown that \ours, together with our parameter
determination method, provides very good performance in comparison to
\lsi\ over a variety of document-topic distributions and
applications-oriented metrics.

\section*{Acknowledgments}
We thank Branimir Boguraev, Roy Byrd, Herb Chong, Jon Kleinberg, Alan
Marwick, Mary Neff, John Prager, Edward So, Charlie Van Loan, and
Steve Vavasis for many useful discussions, and the anonymous reviewers
for their helpful comments.  Portions of this work were done while the
first author was visiting IBM T. J. Watson Research Center.  This
paper is based upon work supported in part by the National Science
Foundation under ITR/IM grant IIS-0081334. Any opinions, findings, and
conclusions or recommendations expressed above are those of the
authors and do not necessarily reflect the views of the National
Science Foundation.

\bibliographystyle{fullname}

\bibliography{../j00m1}

\setcounter{section}{0}
\renewcommand{\thesection}{\Alph{section}}

\bigskip

\section{Appendix}

We use the following perturbation result on singular values, which is
a slight rewriting of Corollary 8.6.2 in \newcite{Golub+VanLoan:96a}.
\begin{theorem}
\label{thm:perturb}
Let $\mv{E} = \mv{X_1} - \mv{X_2}$, where $\mv{X_1}, \mv{X_2} \in
\Re^{r \times s}$ and 
$r \ge s$,  
and for $1 \leq i \leq s$, let
$\sigma_i^{(1)}$ and $\sigma_i^{(2)}$ denote the $i$th singular values
of $\mv{X_1}$ and $\mv{X_2}$, respectively.  Then, \\$| \sigma_i^{(1)}
- \sigma_i^{(2)}| \leq \nTwo{\mv{E}} \leq \normF{\mv{E}}.$
\end{theorem}

\noindent {\bf Notational conventions} ~ If we refer to a singular
value $\sigma_j$ of a matrix $\mv{Z}$ with rank less than $j$, it is
understood that $\sigma_j= 0$. (This corresponds to the ``zero-padded''
version of the SVD that is also commonly used; we
presented the non-padded version above for conceptual clarity.)

Throughout, $\sv_i$ denotes the $i$th singular value 
of $\mProj{opt}(\mAhat)$. 

\subsection{Proof of Theorem \ref{theorem_population}}

Let $\rho_1, \ldots, \rho_n$ be the singular values of the true
topic-based similarities matrix $\simmatrix$.  For convenience, fix a
particular $i \in \{1, \ldots, n\}$.

Since $\fError(\spaceOpt) = \simmatrix - \mProj{opt}(\mAhat)^T
\mProj{opt}(\mAhat)$, by Theorem \ref{thm:perturb} we have $\sv_i^2
\in \rho_i \pm \sOptErr$.

Next, define the matrix $\simmatrix' \in \Re^{n \times n}$ by
$$\simmatrix'[t_1,t_2] = \sum_{d \in \collection} \rel{t_1}{d}
\rel{t_2}{d}\,,$$ where we define $\rel{t}{d} = 0$ for $t > k$ and $d
\in \collection$.  One
can verify that the singular values of $\simmatrix'$ are the same as
the singular values of $\simmatrix$: $\rho_1, \ldots, \rho_n$.

Now, consider the matrix $\mv{E} \stackrel{def}{=} \simmatrix' - {\rm
diag}(\pop{1}^2, \ldots, \pop{n}^2)$.  We note that the singular
values of ${\rm diag}(\pop{1}^2, \ldots, \pop{n}^2)$ are $\pop{1}^2,
\ldots, \pop{n}^2$.  Furthermore, observe that $\normF{\mv{E}} =
\sTsim$, the topic mingling in the collection.  Therefore, by again
applying Theorem \ref{thm:perturb} we find that $\rho_i \in \pop{i}^2
\pm \sTsim$.

Combining our partial results yields the desired result: 
$$\sv_i^2 \in \pop{i}^2 \pm (\sOptErr + \sTsim)\,.$$ 
\qed

\subsection{Invariant subspace tangent theorem}

The proof of Theorem \ref{theorem_1} is based on the following
simplification of the {\em Davis-Kahan tangent theorem}
\cite{Davis+Kahan:70a} (see also Theorem 3.10 of
\newcite{Stewart+Sun:90a}).
\begin{theorem}
\label{thm:stewart1}
Let $\stewmatrix \in \Re^{r \times r}$ be a symmetric matrix, and let
$\makematrix{\mv{X} & \mv{Y}}$ be an orthogonal matrix, with $\mv{X}
\in \Re^{r \times p}$, so that $\Range{\mv{X}}$ forms an {\em
invariant subspace} of $\stewmatrix$ (i.e.  $\mv{x} \in
\Range{\mv{X}}$ implies $\stewmatrix \mv{x} \in \Range{\mv{X}}$).  For
any matrix $\mvapr{X} \in \Re^{r \times p}$ with orthonormal columns,
we define the {\em residual} matrix $\mv{R}$ of $\mvapr{X}$ as
$$\mv{R} = \stewmatrix \mvapr{X} - (\mvapr{X} \mvapr{X}^T) \stewmatrix
\mvapr{X}\,.$$ 
Suppose the eigenvalues of
$\mvapr{X}^T \stewmatrix \mvapr{X}$ lie in the range $[\alpha,\beta]$ and
that there exists $\delta > 0$ such that the eigenvalues of $\mv{Y}^T
\stewmatrix \mv{Y}$ either all lie in the interval $(-\infty,\alpha -
\delta]$ or are all in $[\beta + \delta, \infty)$.  Then,
$$\nTwo{\tan(\Theta(\Range{\mv{X}}, \Range{\mvapr{X}})} \leq
\frac{\nTwo{\mv{R}}}{\delta}\,. $$

\end{theorem}

\subsection{Proof sketch for Theorem \ref{theorem_1}}

Here, we outline the proof (given in full in \newcite{Ando:thesis}).  The
main idea is to apply Theorem \ref{thm:stewart1} by choosing $\mv{X}$
and $\mvapr{X}$ so that $\Range{\mv{X}} = \spaceLSI$ and
$\Range{\mvapr{X}} = \spaceOpt$, and setting $\stewmatrix = \mAhat
\mAhat^T$, for which the \lsi\ subspace is invariant.

Let $\mA = \oProj{opt}(\mAhat)$, and define $\mAbar = \mAhat - \mA$;
note that $\mA^T \mAbar = \mAbar^T \mA = \mv{0}$.  Let $\sigma_i$
denote the $i$th singular value of $\mAhat$.

First, consider the largest singular value of $\mAbar^T\mAbar $, which
we will denote by $\sOrgErrapprox$.  It can be shown that
$\sOrgErrapprox \in \sOrgErr \pm \sOptErr$,
thus justifying our choice of notation.  
(First show that $\fError(\spaceOpt) =
\fError(\spaceOrg)- (-\mAbar^T \mAbar)$, and then apply Theorem
\ref{thm:perturb} to this equation and observe that the largest
singular value of $\fError(\spaceOrg)$ is $\sOrgErr$.)

Then, it can be shown that $\sigma_{h+1} \le \sqrt{\sOrgErrapprox}$:
apply Theorem \ref{thm:perturb} to the equation $\mAbar^T \mAbar =
\mAhat^T \mAhat - \mA^T \mA$ and then note that $\sv_{h+1} = 0$
because of the dimensionality of $\spaceOpt$.

Now, to apply Theorem \ref{thm:stewart1}, choose the columns of
$\mv{X}$ to be the first $h$ left singular vectors (in order) of
$\mAhat$, and choose the columns of $\mvapr{X}$ to be all $h$ left
singular vectors (in order) of $\mA$.  We set the columns of $\mv{Y}$
to the rest of the $m-h$
 left singular vectors (in order) of the
``zero-padded'' version of the SVD of $\mAhat$.

Some linear algebra reveals that the eigenvalues of $\mv{Y}^T
\stewmatrix \mv{Y}$ are no greater than $\sigma_{h+1}^2$, and the
eigenvalues of $\mvapr{X}^T \stewmatrix \mvapr{X}$ are no smaller than
$\svMin^2$.  Therefore, we set $\delta = \svMin^2 - \sigma_{h+1}^2$.
Note that $\delta \geq \svMin^2 - \sOrgErrapprox$ by above, and so is
positive by assumption.  Hence, Theorem \ref{thm:stewart1} applies.

Finally, it can be shown that $\nTwo{\mv{R}} \leq \svMax \cdot
\sqrt{\sOrgErrapprox}$, which yields the desired result:
\begin{eqnarray*}
\nTwo{\tan(\Theta(\spaceLSI,\spaceOpt))} & \leq & \frac{\svMax \cdot
\sqrt{\sOrgErrapprox}}{\svMin^2 - \sOrgErrapprox} \\
& \leq & \frac{\svMax}{\svMin} \cdot
\frac{\sqrt{\sOrgErrapprox}/\svMin}{1 -
(\sqrt{\sOrgErrapprox}/\svMin)^2}\,.
\end{eqnarray*}
\qed
\end{document}